%% file: didl_arxiv_2025.tex
\documentclass[sigconf]{acmart}

\usepackage{booktabs} 

\usepackage{graphicx}


\usepackage{afterpage,natbib,lipsum}
\usepackage{float}
\usepackage{caption}
\usepackage{textcomp}
\usepackage{mathtools}
\usepackage[ruled]{algorithm2e}
\usepackage{leftidx}

\renewcommand\footnotetextcopyrightpermission[1]{} 
\begin{document}
\title{Hopfield Networks Meet Big Data: A Brain-Inspired Deep Learning Framework for Semantic Data Linking }

\author{Ashwin Viswanathan K}
\affiliation{%
  \institution{Oklahoma State University}
  \country{USA}
}
\email{ashwin.kannan@okstate.edu}


\author{Johnson P Thomas}
\affiliation{%
  \institution{Oklahoma State University}
  \country{USA}
}
\email{jpt@cs.okstate.edu}

\author{Abhimanyu Mukerji}
\affiliation{%
	\institution{Stanford University}
	\country{USA}
}
\email{amukerji@stanford.edu}



\begin{abstract}
The exponential growth of data in recent years has resulted in vast, heterogeneous datasets generated from multiple sources. Big data applications increasingly rely on these datasets to extract knowledge for predictive analytics and decision-making. However, the quality and semantic integrity of data remain critical challenges. In this paper, we propose a brain-inspired distributed cognitive framework that integrates deep learning with Hopfield networks to identify and link semantically related attributes across multiple datasets. Our approach models the dual-hemisphere functionality of the human brain, where the right hemisphere processes and assimilates new information while the left hemisphere retrieves learned representations to establish meaningful associations. The cognitive architecture operates on a MapReduce framework and links datasets stored in the Hadoop Distributed File System (HDFS). By incorporating deep Hopfield networks as an associative memory mechanism, our framework strengthens the recall of frequently co-occurring attributes and dynamically adjusts relationships based on evolving data usage patterns. Experimental results demonstrate that attributes with strong associative imprints in Hopfield memory are reinforced over time, while those with diminishing relevance gradually weaken—a phenomenon analogous to human memory recall and forgetting. This self-optimizing mechanism ensures that linked datasets remain contextually meaningful, improving data disambiguation and overall integration accuracy. Our findings indicate that combining deep Hopfield networks with a distributed cognitive processing paradigm provides a scalable and biologically inspired approach to managing complex data relationships in large-scale environments.
 
\end{abstract}

%
%



\maketitle

\input{samplebody-conf}

\bibliographystyle{ACM-Reference-Format}
\bibliography{sample-bibliography} 

\end{document}

%% file: samplebody-conf.tex
\section{Introduction}
In the past decade, there has been a vast change in the landscape related to data handling methodologies and practices. Various advances in Machine Learning (ML) and Artificial Intelligence (AI) algorithms have lead to a more intuitive understanding of data. Neural Networks modeled after the human brain has seen a rapid development with various models like  Artificial Neural Nets (ANN). Recurrent Neural Nets (RNN) being developed and packaged to solve real world problems. These are found to be very attractive in applications like Image classification, Pattern recognition, Object Identification etc.\. Thus semantic understanding of data gives us a better ability to perceive and discern various attributes or representations of data which helps us to effectively integrate with applications. The rise of \textit{Big Data} has led to the development of distributed big data storage and processing frameworks like Hadoop \cite{hadoop}. Several studies highlight the importance of data preprocessing and the role of similarity identification in improving data quality and integration. In this paper we present a novel way of identifying associations in datasets by modeling a brain inspired architecture to understand and learn the changing behavioral pattern in using data, thus providing similarity relationship between datasets. Interest in developing a cognitive system has sparked renewed research interests giving rise to what is today popularly know as \textit{Deep Learning} \cite{wang2017origin}. Human brain modeling started as the precursor to modern deep learning systems. Genesis to such models started with the step of developing computation techniques simulating the human brain. Scientific paradigms like \textit{Brain Inspired Cognitive Architecture (BICA)} are endeavors started to that effect. Another aspect of cognitive systems is concerned with interpreting biological process like thinking and logical reasoning. This leads to the duality of brain hemispheres \cite{goldberg2006wisdom}. Thinking and learning happens in the right hemisphere, and the left hemisphere processes learned information. This presented us with motivation and scientific evidence to realize a cognitive model capable of functioning similar to the human mind, we embarked on developing our framework which is inspired by the aforementioned principles of cognitive computing. We leverage our model to use the MapReduce framework which further enhances the computational ability. We evaluate our model with experiments that brings out the biological responses of the human mind. Recent research work in BICA \cite{chernavskaya2013architecture, chernavskaya2016natural, chernavskaya2017modelling} consider using Hopfield type of processors to model the hemispherical structure of the brain. Adopting a similar approach, we showcase a novel way of using Hopfield networks to learn and understand similarity relations across datasets. We essentially treat usage information as a pattern that can be learnt by our model. This provided an interesting approach at understanding and deriving data usage information which are then read by Hopfield nets. Hopfield nets \cite{hopfield1982neural} are a form of RNN, the recurrence property stems from the fact that, the neurons or nodes are bi-directional. They also possess content addressable memory which is biologically similar to the human brain in forming associations. Historically Hopfield nets belong to the earliest class of \textit{Deep Learning systems} \cite{wang2017origin}. The associative memory of Hopfield nets is suitable in most applications. Pattern and image recognition are some of the most common applications which use Hopfield networks. In this paper, we present an interesting approach to use Hopfield nets to clean data by modeling it as a pattern recognition task. We use MapReduce thus augmenting its capacity to process voluminous data. In our research, we adapt a data usage tracking approach and integrate it with our proposed cognitive brain model to learn from associations and identify similar datasets.\\
The contributions of our work is two-fold
\begin{itemize}
\item We provide theoretical research and a basis for realizing a biological model following the tenets of cognitive science and workings of biological neurons. We propose a framework which uses the Hopfield network with MapReduce on top as computational units to learn and gather information or knowledge. 
\item To validate our framework, we apply our brain inspired model to learn and identify
semantic links amongst datasets. This is a first step in the process of data cleaning for big data.

\end{itemize}
We present experimental evaluations to prove the validity and correctness of our model. In section \ref{rel_work}, we present related work in cognitive architecture and cleaning of data in big data applications. Section \ref{data_clean} describes how data cleaning is achieved using our model and Section \ref{framework} describes the overview of our brain architecture, and implementation details follows in subsequent sections. Evaluation of our model is shown in Section \ref{exp} and validation is shown in Section \ref{results}. Conclusion and future scope is shown in Section \ref{concl}.

\section{Related Work}\label{rel_work}
Our architecture draws upon research conducted in BICA which focuses on creating a computational model equivalent to a human mind. Within BICA, a principle called the \textit{Dynamic Theory of Information (DTI)} was proposed by \cite{chernavskaya2013architecture} which gives a formal description of computationally modeling the components of human mind (i.e) \textit{memory, thought and emotions}. The concept of \textit{Dynamic Formal Neuron(DFN)} proposed by \cite{chernavskaya2016natural} talks in detail of building a neuron processor using \textit{Hopfield} \cite{hopfield1982neural} and \textit{Grossberg} \cite{grossberg1982studies, grossberg1987adaptive} processors. The concept of brain hemispherical duality was explored in \cite{chernavskaya2017modelling}. This provided a premise to model a framework analogous to the human brain. Further studies by \cite{laird2008extending} indicate the means to realize cognitive structures. Based on the above research and aligning closely with the works of \cite{chernavskaya2013architecture} and \cite{chernavskaya2016natural}, we chose the Hopfield Network to model our architecture inspired by biological neural networks. There has been a renewed interest in research about creating structures similar to the neurons found in the brain. ANN's \cite{gupta2013artificial, wang2003artificial} emerged as the foremost computing systems whose functioning resembled the human mind. Principles of self-organization \cite{kohonen1990self} which uncovers semantic relationship in sentences are some of the features present in ANN. Based on the above research works, our model can be applied to identify associations or similarity in datasets. This problem falls into the category of pattern recognition. Identifying similarity/associations among datasets helps in efficient cleansing. Similarity and associations between datasets have been extensively leveraged for identifying sensitive data items and data cleaning. Existing research explores the integration of data usage patterns and contextual information to enhance the identification of semantic associations between data items. One such approach employs linear weighted equations for context-based similarity assessment and utilizes Markov’s clustering algorithm \cite{van2001graph} for semantic grouping. However, a major limitation of this methodology lies in the estimation of optimal equation weights and the identification of appropriate constants for Markov’s clustering, which can significantly impact clustering accuracy and computational efficiency \cite{van2001graph}. As the work presents sufficient scope for pattern recognition, we rewire a section of the architecture that is, identifying semantically related datasets found in the above work to suit the brain inspired framework proposed in this paper. We also evaluate the behavior of our neural nets by making use of usage patterns similar to those found in the real world and present the statistical approach for defining
accuracy in identifying similarity.

\section{Data Cleaning}\label{data_clean}
In this paper, we focus on using our framework to identify similar data-items across datasets, which is then applied to clean data. Data usage information is used to identify semantic similarities \cite{niemann2011usage}. Semantic similarity gives a measure of the usefulness of the data-item as they are captured from previous usage history information. In this paper we use an item based method \cite{sarwar2001item} to identify similar datasets. Usefulness of data is indicated by how frequent it is accessed \cite{niemann2011usage}. Hence, we consider similarity as a suitable measure to ascertain data usefulness. In our research, we capture the usage characteristics as a vector class of patterns, consisting of ($-1, +1$) values, where $+1$ indicates relatedness and $-1$ as not related. This is then used to train our brain model which has an unsupervised learning procedure. On repeated training, our model learns the patterns which correspond to the most frequently used dataset. This enables to identify similarity of datasets when the model is asked to predict similarly related data-items when given a data usage class of patterns.

\section{Biological Neuron Framework for Semantic Identification}\label{framework}

\subsection{Overview of Brain Based Architecture}
The foundation of our model is based on the emergent scientific paradigm BICA. It aims to design, study and implement human-level cognitive architectures \cite{laird2008extending}. Recent works on BICA \cite{chernavskaya2013architecture, chernavskaya2016natural} proposed a \textit{Natural-Constructive Approach (NCA)} to model a cognitive system. The premise for developing our model is based on the above works and \cite{goldberg2006wisdom}  and the hemispherical functions of the brain and concept of thinking process postulated in DTI \cite{chernavskaya2017modelling}. Our framework, similar to NCA consists of two sub-models namely the right and left hemispheric layers. Information gathering and learning occurs in the right hemisphere and processing the learned information occurs on the left.
\begin{figure}[h]
	\centering
	\captionsetup{justification=centering,margin=2cm}
	\includegraphics[scale= 0.5]{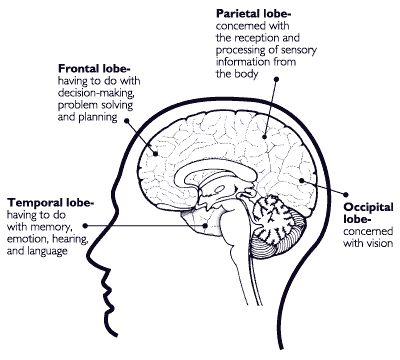}
	\caption[Human Brain Representation]{Human Brain Representation\protect\footnotemark}
	\label{fig:br_lobes}
\end{figure}
Our proposed model aligns with the biological aspects of the human brain, in that, we follow the learning and recalling ability of the human brain when presented with external stimuli. Cerebral cortex and sub-cortical structures of the brain are responsible for cognitive, emotional and memorization processes \cite{stirling2010introducing}. Figure \ref{fig:br_lobes} depicts the functions of different lobes of the brain. Table \ref{tab:table1} presents lobes of the brain and its computational interpretation.
\begin{table}[h!]
	\centering
	\caption{Brain Lobes and Computer Structures}
	\label{tab:table1}
	\begin{tabular}{lll}
		\toprule
		Brain Structure    &    Function   &   Interpretation\\
		\midrule
		Occipital Lobe & Vision  & Any multimedia data.  \\
		Parietal Lobe  & Motor  & Trajectory data points. \\
		Temporal Lobe  & Auditory  & Text perception and reproduction. \\
		Frontal Lobe   & Speech & Information propagation and relaying.\\
		\bottomrule
	\end{tabular}
\end{table}
In this paper, we model the functionality of temporal lobe found in right side of the brain from figure \ref{fig:br_lobes} by learning and retrieving representations of data usage patterns. We describe the building blocks of our model using Hopfield Nets which belong to the class of RNN. Then we discuss the learning rules and associations formed when exposed to information. Finally we talk about generated experiments and results. Deep Belief Network library called \textit{Neupy} \cite{DBNAlbert} is used to implement our model. 

\subsection{Brain Inspired Architecture for Identifying Associations}
\begin{figure*}[h]
	\centering
	 \captionbox[Caption]{Brain Model Overview. An input class pattern (vectors) is first fed to the layer of Hopfield neurons which forms the temporal lobe region in the right hemisphere (RH). Normalized vectors are stored in weight matrix (Hippocampus). Noisy patterns sent to Wernicke region in left hemisphere (LH) processes and predicts the output (Broca's region). The LH receives learned information from RH. \label{fig:brain_arch}}{
	\includegraphics[scale=1]{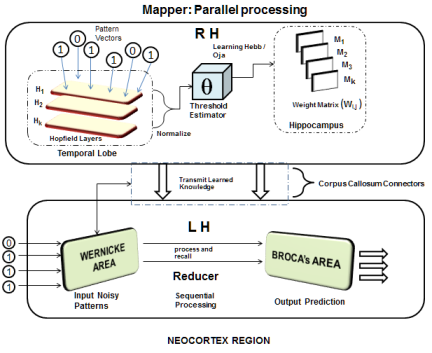}
}
	\label{brain_arch}
\end{figure*}

\footnotetext{National Institute on Drug Abuse (1997) Mind Over Matter: The Brain's Response to Drugs, Teacher's Guide}

Our proposed architecture (see figure \ref{fig:brain_arch}) draws on the aforementioned concepts of cognitive modeling. Neural physiology \cite{solso2013cognitive} and \cite{bianki1984parallel} tells us that information is processed and learnt \cite{goldberg2009new} in a parallel manner by the right hemisphere. In our model, as in the brain, learning of new information occurs in parallel on the right side of the brain. This is achieved by using Hopfield nets in tandem with the computational power of \textit{MapReduce} and \textit{HDFS} to process enormous volume of data. MapReduce and HDFS is a storage and parallel processing model for big data \cite{lee2012parallel, ekanayake2008mapreduce}. Our  model is analogous to the \textit{Right Temporal Lobe} shown in figure \ref{fig:brain_arch}. The noise/information is perceived by layer of Hopfield networks and the interconnections serves as a conduit for transferring learned information. Internal memory of the Hopfield neurons is similar to the hippocampus of human brain. It is involved in the storage of long-term memory, which includes all past knowledge and experiences. Data vectors with values of either $-1$ or $+1$ are learned and stored by the Hopfield nets. Sequential Information processing is done on the left side of the brain.
\subsubsection{\textbf{Right Hemisphere (RH):}} Learning and perception of knowledge occurs in the right side of the brain. In our model (as shown in figure \ref{fig:brain_arch}), the layer of Hopfield nets presides on the right side which complements the biological brain structure. An instance of our system is represented by a class of vectors $(a_1 , a_2 , a_3 , ... , a_k)$ where each of $a_1$ to $a_k$  is either $+1$ or $-1$ after passing through \textit{Threshold Estimator}. The weight matrix $W_{i,j}$ which stores the memorized pattern is analogous to the \textit{Hippocampus} of brain. This region contains the \textit{Mapper} process which parallelizes the learning process. 
\begin{algorithm}[t]
	\SetAlgoNoLine
	\KwIn{Data usage information pattern matrix ($k \times k$)}
	\KwOut{Induvidual chunks of trained memory matrices $w_{i,j}$}
		$a \gets Read~ matrix ~of ~patterns ~from ~file$
		$norm_a = EucNorm(a)$ \tcp*{normalize}
		$bin_a = TE(norm_a)$ \tcp*{update to $+1$ or $-1$}
		$W_{ij} = HopfieldLearn(bin_a)$ \tcp*{Train patterns}
		
		\For{each $w_{i,j}$ ~ term ~in ~$W_{i,j}$} {
			$EMIT_{intermediate}(w , count)$
		}

	\caption{Parallelizing Right Hemisphere (Mapper)}
	\label{alg:alg_rh}
\end{algorithm}
The frequency of data usage patterns is generated for a set of $k$ datasets which is then normalized using standard euclidean norm \footnote{http://slideplayer.com/slide/5057707/} given by the below equation \ref{euc_norm}. This bounds the values between $0$ and $1$.
The algorithm for this layer is given above.
\begin{equation}\label{euc_norm}
\lVert A \rVert_{e} = \sqrt{\sum_{i = 1}^{n} \sum_{j = 1}^{n}\left|{a_{i,j}}\right|^{2}}
\end{equation}
\subsubsection{\textbf{Threshold Estimator (TE):}}\label{te} Once the frequency of usage has been captured and normalized using equation \ref{euc_norm}, the normalized usage matrix are received by the \textit{Threshold Estimator}. For any $k_n$ datasets, suppose there is high frequency of usage between $k_i$ and $k_j$, then these states must be active (i.e) a value of $+1$ is assigned in the symmetric memory matrix $W_{i,j}$ of the Hopfield nets. In this paper we refer to the \textit{actual learned pattern} as class of vectors which already have defined binary states of $+1$ or $-1$ based on the recurrent data usage rules. These class of vectors are then used as a basis by the \textit{threshold estimator} to converge to the correct pattern. The point of convergence varies for each user of datasets due to change in frequency of usage. 
\subsubsection{\textbf{Modeling the Hopfield neurons in RH}}:
\label{model_hem_layers}
Hopfield nets \cite{hopfield1982neural}\cite{wiki:Hopfield_network} provides distributed memory and intuitively mimics the workings of human brain. They undergo an unsupervised learning procedure and possess binary states that is, $+1$ or $-1$. They are represented as a pair of \textit{\textbf{$i$}} and \textit{\textbf{$j$}} neurons. Neuron interconnections is described by the connectivity weight matrix \textit{\textbf{$W_{i,j}$}}. They are densely interconnected and the weight matrix is symmetric with the diagonals having a value of $0$. Hopfield nets belong to the class of \textit{\textbf{McCulloch and Pitts}} neurons \cite{mcculloch1990logical}. For every \textit{$k$} class of patterns we require \textit{$k$} Hopfield neurons. There are two modes for updating the internal states of $W_{i,j}$ that is, Synchronous and Asynchronous. Refer to section \ref{upd_modes} for more details on the update modes. Our paper follows the \textit{asynchronous} mode due to its similarity with biological memory. 

\subsubsection{\textbf{Left Hemisphere (LH):}} Reception of information and processing the learned knowledge is carried in the left side of the brain. In this region, information is processed in a sequential manner \cite{bianki1984parallel}. This layer uses the \textit{Reducer} process of the MapReduce framework to combine the individual chunks of memory information from the RH. Algorithm for this layer is presented below.

\begin{algorithm}[t]
	\SetAlgoNoLine
	\KwIn{Input patterns stored in distributed Hadoop cache}
	\KwOut{Output pattern containing similar data associations}

	\For{each $w_{i}$ ~ in ~ $w$} {
		$W_{i,j} ~+= w_{i}$
	}

	\For{each ~test ~pattern ~$p_{t}$ ~from ~cache} {
		
		$predict_{pattern} \gets RHHopfieldRecall(W_{i,j} , p_{t})$
	}
		
	\caption{Left Hemisphere Prediction (Reducer)}
	\label{alg:alg_lh}
\end{algorithm}

Once the network is trained with data vectors on the RH, we can question the model to identify data similarity across given $n$ datasets. This process of recalling is conceptually equivalent to the \textit{Wernicke Area} present on the left side of the brain which is responsible for speech comprehension. The output region of the network is similar to the \textit{Brocka's Area} (as shown in figure \ref{fig:brain_arch}) which is responsible for speech production.\\
From the functions of both the layers, we see a inherent equivalence between biological neurons and artificial Hopfield nets implemented in our model.


\subsection{Palimpsest Memory}
Connection weights among neurons store the networks memory. Hopfield nets \cite{hopfield1982neural} recollect previously stored patterns when presented with noisy input. In our model, we implement palimpsest or forgetful rules for learning where the network being exposed to patterns gradually loses its ability to recall earlier or less repetitive patterns. Below we discuss the learning rules applied in our model.

\subsubsection{\textbf{Hebbs and Oja's Learning rule:}}
One of the foremost theories of learning was proposed by Donald Hebb in \cite{hebb2005organization}. Hebbian theory belongs to unsupervised learning. When nodes fire simultaneously, their connection strength increases. From \cite{hebb2005organization} and \cite{wiki:Hopfield_network}, $W_{i,j}$ is given by
\begin{equation}\label{hebb_rule}
W_{i,j} = \frac{1}{n}\sum_{p = 1}^{k}x_{i}^{p}x_{j}^{p}
\end{equation} \\
\textit{Erkki Oja} introduced a new rule modifying hebb's rule known as Oja's rule \cite{oja1982simplified}. It stabilizes hebb's rule by introducing a normalization factor. Refer to sections \ref{apd_learning} for further details.
\subsection{Experiments} \label{exp}
\subsubsection{\textbf{Dataset:}} We conduct our experiments by capturing data usage information for a given $k$ datasets. In previous research, a data usage tracking component was employed to capture usage behavior. In our work, we refine this approach by simulating and generating $k$ datasets and $p$ combinations of usage patterns to enhance association learning and dataset similarity identification. Since our model runs on top of MapReduce, the values of $k$ and $p$ can be sufficiently huge leveraging the capacity of HDFS. The usage pattern is generated  with different values of $k$ and $p$ to evaluate the cognitive learning of our framework.

\subsubsection{\textbf{Evaluation:}}
For any $k$ datasets, a relation between $k_{i}$ and $k_{j}$ is said to exist when they are frequently used and their usage count is large enough to cross the threshold limit set by \textit{TE}. Thus a value of $1$ between $k_{i}$ and $k_{j}$ is said to indicate an association. Since the weight matrix $W_{i,j}$ is bi-directional $k_{i,j} = k_{j,i}$. Evaluations were conducted by storing $k \times k$ matrix of patterns in the Hopfield network where $k$ represents the datasets. Each usage pattern is treated as a row of vectors $a_{i,k}$ that is, each user's data usage information is captured as a single row in the $k \times k$ matrix. A single row can be represented as $a_{i,k} = [a_{1,1}, a_{1,2},  .... , a_{1,k}]$. Thus for any $k \times k$ dimensions, the matrix of usage information is represented by\\

$A_{k,k} = 
\begin{bmatrix}
	a_{1,1} & a_{1,2} & \cdots & a_{1,k} \\
	a_{2,1} & a_{2,2} & \cdots & a_{2,k} \\
	a_{3,1} & a_{3,2} & \cdots & a_{3,k} \\
	\vdots  & \vdots  & \ddots & \vdots  \\
	a_{k,1} & a_{k,2} & \cdots & a_{k,k} 
\end{bmatrix}$ \\

Let the patterns generated be represented as $p = (p_{1} , p_{2} , p_{3} , ... , p_{n})$. Each of $p_{i}$ may contain varying number of relationship links, for example $k_1k_2k_5$ in $k_{n}$ datasets indicates an association link between $k_1 \leftrightarrow k_2$, $k_1 \leftrightarrow k_5$ and $k_2 \leftrightarrow k_5$ such that their corresponding entries would have a value of $1$. In the previous example, the number of relationship links was 3, in general there could be $n$ related links amongst datasets. We performed the experiments by training the right hemisphere with $p_{i}$ patterns initially where $i$ could be any value. To illustrate the concept of learning associations and exhibiting forgetfulness and recall, let us consider the initial value of $i$ to be $1$. RH includes the \textit{Mapper} which processes in parallel and splits the workloads into intermediate records. Internally for each pattern a corresponding weight matrix $W_{i,j}$ is produced by the mapper. Now the memory of Hopfield networks contain pattern $p_{i}$ which is stored in the network. Let $\alpha_{0}$ denote the instance at which $p_{i = 1}$ is trained. Test patterns can then be fed to the LH for predicting similarity of datasets. For a test pattern $p_{t_{i = 1}}$, the LH which includes the Reducer process, combines the individual weight matrices and then generates the closest matching pattern. The resulting pattern $r_{i = 1}$ is then used to calculate the number of bits of relationship links lost $\beta$ and new associations added $\gamma$. We first retrieve the indices from both the test $p_{t_{i = 1}}$ and result pattern $r_{i}$. Let $\zeta$ denote the index positions having associations. Then $\zeta$ can be written as:

\begin{equation} \label{zeta}
\zeta_{i,j} = (i,j) ~ \mbox{if $X_{i,j} = 1$}
\end{equation}

where $X_{i,j}$ $\in$ $(p_{t_{ij}}, r_{ij}, p_{ij})$.\\
From equation \ref{zeta} we get a tuple of $(i , j)$ values for each of $(p_{t_{ij}}, r_{ij}, p_{ij})$. We can map the index positions to retrieve the dataset labels. This gives the \textnumero ~dataset relations in each class of patterns. \\
Let $P_{T}$ denote the set of data associations found from equation \ref{zeta} when $X_{i,j}$ = $p_{t_{ij}}$  and $R$ be the set of associations from equation \ref{zeta} when $X_{i,j}$ = $r_{ij}$ in the result pattern. We calculate $\beta$ as:\\

\begin{equation} \label{eq:beta}
\beta = P_{T} \cap \mathrm{c}(R)
\end{equation}\\
In equation \ref{eq:beta}, $\mathrm{c}(R)$ denotes the $complment$ of result pattern $R$

Similarly let $P$ contain the set of original associations present in Hopfield network, then $\gamma$ is calculated as: \\

\begin{equation}\label{eq:gamma}
\gamma = R \cap \mathrm{c}(P)
\end{equation}\\
In equation \ref{eq:gamma}, $\mathrm{c}(P)$ denotes the $complment$ of original pattern $P$. At instance $\alpha_{0}$, from equation's \ref{eq:beta} and \ref{eq:gamma}, we also calculate the \textit{cosine similarity} between $p_{t_{i}}$ and $p_{i}$, and  between $r{i}$ and $p_{i}$. This gives us a similarity score between the patterns found in Hopfield's memory to that of the test and result patterns. Higher scores indicate similarity to patterns stored in Hopfield network.
\section{Results} \label{results}
When $i = 1$, the value of $\beta$ and $\gamma$ is $0$ which indicates $100$\% recovery accuracy as no patterns are lost and no new associations are added. The result pattern $r_{i}$ has bits turned on ($+1$) for association links, which the model deduces as having similarity. This happens due to strengthening of the networks memory by storing similar usage information repeatedly. At instance $\alpha_1$, increasing the value of $i$ to $i +1$, we store more patterns ($p_{i + 1}$) which are not similar to the initial trained pattern $p_{i = 1}$ in the Hopfield network. By having the same test pattern $p_{t_{i = 1}}$ to predict, we observe that the values of both $\beta$ and $\gamma$ increase gradually thus diminishing the recovery accuracy. Higher value in $\beta$ indicates the forgetfulness experienced by the model. It also highlights the \textit{change in usage information behavior}. The high value in $\gamma$ denotes forming new associations which are closer to the patterns being currently stored in Hopfield's memory from instance $\alpha_1$ onwards. It also means, that associations found in the test pattern has weaker relationships and is less likely to be frequently. This is a measure of the framework's self-learning capacity. This behavior directly relates to the biological functioning of the human memory and adapting based on dynamic changes in usage information. The results and calculated metrics from these evaluations shows that our proposed model is able to identify similar datasets based on usage data.

\section{Conclusions} \label{concl}
In this paper, we have introduced a cognitive and distributed framework for identifying similarity associations among datasets. Our model leverages MapReduce to process vast volumes of data efficiently, enabling scalable and parallel computations. Inspired by the dual processing mechanisms of the cerebral hemispheres in the human brain, our architecture incorporates neural associative memory for effective information retrieval and pattern recognition. The right hemisphere of our framework integrates modern Hopfield networks, a class of energy-based models renowned for their associative memory capabilities, to retrieve and encode similarity relations. The resurgence of Hopfield networks in deep learning—particularly their continuous-state generalization and capacity for high-dimensional memory storage—provides a biologically plausible mechanism for handling large-scale data associations. This design mirrors the right temporal lobe of the human brain, where memory recall and associative cognition predominantly occur. To enable efficient parallel computation of information usage, we incorporate the Mapper process, facilitating distributed similarity detection. The left hemisphere serves as the predictive reasoning unit, analogous to the Wernicke and Broca regions, which process and synthesize linguistic and conceptual knowledge. Here, the Reducer module integrates fragmented associative patterns from the right hemisphere and generates structured predictions regarding data similarity. This design aligns with deep learning methodologies where hierarchical feature integration plays a pivotal role in contextual understanding. Our evaluation metrics, including measures for identifying lost associations ($\beta$) and predicting potential semantic relations ($\gamma$), demonstrate the efficacy of our model in cognitive data processing tasks. Beyond traditional applications in data cleaning, our framework is well-suited for contextual similarity discovery in large-scale datasets, an increasingly critical challenge in fields such as automated knowledge graph construction, information retrieval, and unsupervised learning for semantic inference. Future research will focus on extending our framework to leverage continuous Hopfield networks and transformer-based architectures for enhanced long-range dependencies and dynamic adaptation in real-time data environments.

\appendix
\section{Appendix}
In this appendix, we look at Hopfield Network's equations and learning rules.
\subsection{Hopfield Networks}
The following holds true regarding the weight matrix:
\begin{itemize}
	\item \textit{\textbf{$W_{i,i}$}} = 0 $\forall i$
	\item \textit{\textbf{$W_{i,j}$}} = \textit{\textbf{$W_{j,i}$}} $\forall i, j$
\end{itemize}
\subsubsection{Update Modes}\label{upd_modes}
From \cite{wiki:Hopfield_network} the update rule of Hopfield process for state $s_i$ is given by:
\begin{equation} \label{Hop}
s_i:=
\left\{
\begin{array}{ll}
1 & \mbox{if $w_i + \sum_{j=1}^n w_{ij} s_j(t) \geq \theta$} \\
-1 & \mbox{otherwise.}
\end{array}
\right.
\end{equation}
Here $\theta$ is derived from \textit{TE} described in section \ref{te}.
In \textit{synchronous} mode, each of the $k$ neurons simultaneously evaluates and updates its state. In \textit{asynchronous} mode, neurons are chosen at random and updated one at a time . 

\subsubsection{Learning Rules}\label{apd_learning}
Hebbian principle rests on the fact that when an axon of neuron $i$ repeatedly takes part in firing another neuron $j$, then the strength of connection between $i$ and $j$ increases. This rule is both local and incremental. it is implemented in the following manner, when learning $k$ binary patterns:
\begin{equation}
\Delta w_{i,j} = \alpha x_{j} \ast y_i
\end{equation}
where $\epsilon_{i}^{p}$ represents bit $i$ from vector pattern $p$. Capacity of this rule is $0.14k$ where $k$ is number of distinct patterns. Gradual decay of memory is experienced over a period of time. Continued exposure to similar patterns results in a stronger memory retention while recalling.\\
The main idea of \textit{Oja's} rule was to normalize each neuron to a value of $1$ \cite{szandala2015comparison}. From \cite{szandala2015comparison}, we present the equation for $W_{i,j}$
\begin{equation}
|W_i| = 1
\end{equation} 
To achieve this, we add a $V$ factor such that:
\begin{equation}
V = \sum_{k=1}^{m}W_{i,j} \ast x_j^{k}
\end{equation}
The weight matrix is updated by:
\begin{equation}
W_{i,j}^{k + 1} = W_{i,j}^{k} + u \ast V \ast (x_j^{k+1} - V \ast W_{i,j} ^{k})
\end{equation}
